%% file: paper.tex
\documentclass{article}

\usepackage[final]{corl_2020} 
\usepackage{times}
\usepackage[numbers]{natbib}
\usepackage{multicol}
\usepackage{graphicx}
\usepackage{multirow}
\usepackage{amsmath,amssymb} 
 
\usepackage{color}
\usepackage[dvipsnames]{xcolor}
\usepackage{hyperref}
\usepackage{tikz}
\usepackage[utf8]{inputenc}
\usepackage{pgfplots}
\usepackage{subfig}
\usepackage{booktabs}
\usepackage{adjustbox} 
\usepackage{breakcites}
\usepackage{url}
\usepackage{siunitx}  
\usepackage{pifont}
\usepackage{amsmath}
\usepackage[title]{appendix}
\DeclareCaptionLabelFormat{andtable}{#1~#2  \&  \tablename~\thetable}

\usepackage[acronym,toc]{glossaries} 

\newcommand{\cmark}{\ding{51}}%
\newcommand{\xmark}{\ding{55}}%

\title{Self-Supervised Object-in-Gripper Segmentation from Robotic Motions}
\author{
  Wout Boerdijk$^1\qquad$ Martin Sundermeyer$^1\qquad$ Maximilian Durner$^1 \qquad$  Rudolph Triebel$^{1,2}$ \vspace{0.1cm}\\
  \begin{tabular}{c p{0.3cm} c}
$^1$Institute of Robotics and Mechatronics & &  $^2$ Department of Computer Science \tabularnewline
German Aerospace Center (DLR) & &Technical University of Munich (TUM)  \\
    82234 Wessling, Germany&& 85748 Garching, Germany\\
    \texttt{Firstname.Lastname(at)dlr.de}&&
\end{tabular}
}

\begin{document}
\maketitle


\begin{abstract}
Accurate object segmentation is a crucial task in the context of robotic manipulation.
However, creating sufficient annotated training data for neural networks is particularly time consuming and often requires manual labeling.
To this end, we 
propose a
simple, yet robust
solution
for learning to segment unknown objects grasped by a robot.
Specifically, we exploit motion and temporal cues in RGB video sequences.
Using optical flow estimation we first learn to predict segmentation masks of our given manipulator. 
Then, these annotations are used in combination with motion cues to automatically
distinguish between background, manipulator and unknown, grasped object. 
In contrast to existing systems our approach is fully self-supervised and independent of precise camera calibration, 3D models or potentially imperfect depth data.
We perform a thorough comparison with alternative baselines and approaches from literature. 
The object masks and views are shown to be suitable training data for segmentation networks that generalize to novel environments and also allow for watertight 3D reconstruction. 
\end{abstract}



\section{Introduction}
Despite the major advancements recently made in robotic perception, in practice there are still a lot of challenges to be solved. Two of the most prominent are (1) the lack of annotated training data for variable recognition tasks and (2) domain gaps originating from varying environments and sensors. 
Techniques like transfer-, semi- and meta- learning  can lower the amount of necessary annotations but they do not abstain from it. Another approach is to leverage synthetic data while ensuring generalization to real data through photo-realistic rendering, 
domain randomization or domain adaptation (e.g. ~\cite{hodan2019photorealistic, sundermeyer2018implicit}). 
However, creating the required 3D content is a laborious process and real labeled sensor recordings from environments relevant for execution often yield better results.

\begin{figure}[!t]
	\centering
	\captionsetup{width=\linewidth}
	\includegraphics[width=\linewidth]{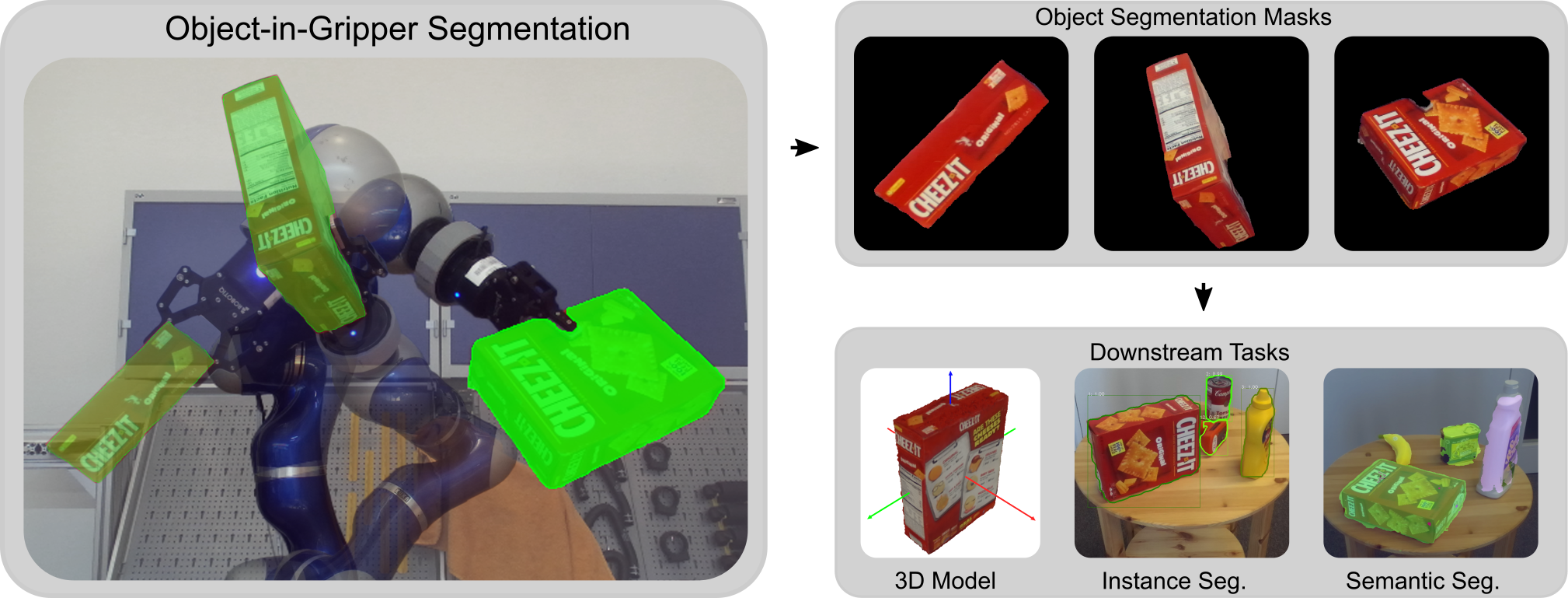}
	\caption{Our self-supervised method is able to segment unknown grasped objects from robotic motions. The resulting annotation masks can be applied to several downstream tasks such as 3D modeling or as training data for instance or semantic segmentation.}
	\label{fig:front}
\end{figure}

In this paper we investigate a different approach: Robots can concurrently observe and interact with objects similar to the way human infants 
examine them
\cite{intro_infantslearning2014lobo}.
Concretely, we perform motions of a grasped object in front of a camera and analyze the differences over sequences of images. 
While humans accomplish discerning objects in motion 
with specific velocity-sensitive cells 
\cite{of_visualperception1974Nakayama}, 
we rely on purpose-made Convolutional Neural Networks (CNNs) like LiteFlowNet \cite{of_liteflownet2018hui}. 
These correlate feature maps of subsequent video frames to predict dense optical flow fields. 
This low-level vision task generalizes well to unseen environments and can be trained efficiently in simulation. 
To differentiate the resulting joint object-robot mask we refrain from direct object-agnostic segmentation and instead employ a semantic segmentation network to identify the manipulator from static frames in a self-supervised manner. 
During inference, the trained model removes the robotic arm from the optical flow field to segment the unknown grasped object. 
Thereby, we automatically obtain accurate masks of any graspable object from any possible viewpoint. These can be used in various downstream tasks like self-supervised semantic or instance segmentation and even full 3D object reconstruction from a grasped object (see Figure~\ref{fig:front}).

Our method has major advantages. First, it is independent of precise camera calibration, allowing to place an RGB camera loosely pointing towards any robotic manipulator. Second, it does not rely on 3D robot or object models for segmentation. 
The simple setup enables incremental data collection in varying environments.
For the mask generation we only assume that objects are graspable and that the background is mostly static. There are no assumptions on strict object rigidity or texture. 
Finally, our approach outperforms competing self-supervised methods yielding accurate segmentation performance in novel scenes. 

\section{Related Work}


We review existing approaches that induce motion in a scene in order to perform object segmentation. For an extensive analysis on interactive perception we refer the reader to \cite{perception_ipsurvey2017bohg}.


\subsection{Object Segmentation from Motion}

Motion as an additional cue is a strong enhancement for unknown object segmentation and has widely been investigated.
A particularly na\"ive form thereof is change detection, where subtraction between a pair of images yields the object in motion 
\cite{cd_definition1989singh}.
More sophisticated methods model the statistical representation of a scene background with, for instance, (multiple) Gaussian distribution(s) 
or kernel density estimation \cite{kenney2009interactive, rw_backgroundsubkde2002elgammal}.
Recently, CNNs have also been applied with promising 
results, and we refer the reader to \cite{bouwmans2018deep} for an overview.

Spatio-temporal clustering for object segmentation has first been applied by 
\cite{shi1998motionsegm}. 
The main idea
is to group pixels based on similar motion behavior obtained by tracking their motion with optical flow over a sequence. 
Optical flow is a vector field describing the displacement directions and velocities of each pixel between two subsequent images \cite{of_of1981Horn}. This underconstrained problem can be solved either by means of energy-based optimization or by employing a learning-based approach~\cite{zhigang2019survey}.
\citet{fragkiadaki2015learning} apply a CNN for object motion detection which is forwarded to a clustering method to obtain object masks.
In~\cite{bideau2018best}, classical geometric knowledge via optical flow is combined hierarchically with semantic segmentation obtained by a CNN.
Recently, two works propose to merge appearance (RGB) and motion (optical flow) information either in a two-stream CNN \cite{dave2019towards} or in a recurrent neural network \cite{xie2018objdiscovery}.
\citet{robot_motionsegmentation2018shao} train on synthetic RGB-D frames to estimate object masks as well as a dense 3D motion field, also called scene flow.

These earlier works utilize motion for either binary foreground segmentation or as cue for multi-class supervised segmentation.
Instead, we use optical flow to separate two unknown moving instances from each other without requiring any manual annotations.


\subsection{Robot-based Object Segmentation}
In most cases, robots face a static scene which typically does not move by itself.
A robot however can enhance visual perception by creating helpful motion in an environment, an idea which has initially been suggested by \cite{robot_motionsegmentation2003fitzpatrick}.
Other works follow this approach by inducing object motion via non-prehensile actions 
\cite{kenney2009interactive, robot_motionsegmentation2014schiebener}.
In \cite{schiebener2011segandlearning}, generated object hypotheses are verified by pushing, and emerging feature motions are then clustered to object masks.
The restriction of textured objects is removed in the subsequent work by applying color annotated stereo data for the hypotheses generation ~\cite{robot_motionsegmentation2014schiebener}.
These probabilities are updated based on optical flow and geometric change detection induced by robotic manipulations.
\citet{pathak2018learning} propose an active agent who learns to segment by trying to pick-and-place objects from one location to another.
By observing eventual pixel changes within the grasp region object hypotheses are enforced.

Merely pushing, picking and placing objects falls short in generating diverse object views, a characteristic known to be beneficial for the generalization ability of neural networks and essential for 3d object reconstruction.
However, robotic manipulators can also deterministically control the motion of objects in their gripper and thus efficiently generate vast and diverse viewpoints.
Deliberately generating trajectories of objects emulates the infant learning process of perceiving objects 
\cite{of_visualperception1974Nakayama}.
Furthermore, the progress in unknown object grasping in recent years \cite{rw_visiongraspingsurvey2019du} encourages us to explore this path to increase robot autonomy.

Most recently, Rocha et al. \cite{rocha2019self} utilize the kinematic model of a robot for automatic object segmentation in a surgical setting. Given a robot-camera calibration together with a robot model, they project the latter onto the image plane. By iteratively optimizing a GrabCut-based
cost function on a set of images followed by projecting the robot arm given the respective kinematics, they obtain segmentation labels. These are used to train a FCN, and a post-processing by a CRF results in tool segmentation.
The subsequent work of \cite{florence2019self} builds upon this and differs from the aforementioned method by using joint locations in combination with a depth sensor instead of a kinematic robot model projected onto the image frames. 

In this work, we propose a novel pipeline to automatically generate training data in a self-supervised setting. Little constraints on the setup and known deep learning components ensure simple practical appliance.
In contrast to existing methods
we solely use a single RGB sequence for all tasks and do not incorporate depth imagery which is susceptible to errors.
Furthermore, we do not require precise camera calibration or registration of object key points in the camera frame.
Finally, we explicitly optimize the trajectory in order to maximize the object views.

\section{Method}

\label{sec:method}

\begin{figure*}[ht]
	\centering
	\captionsetup{width=\linewidth}
	\includegraphics[width=0.9\linewidth]{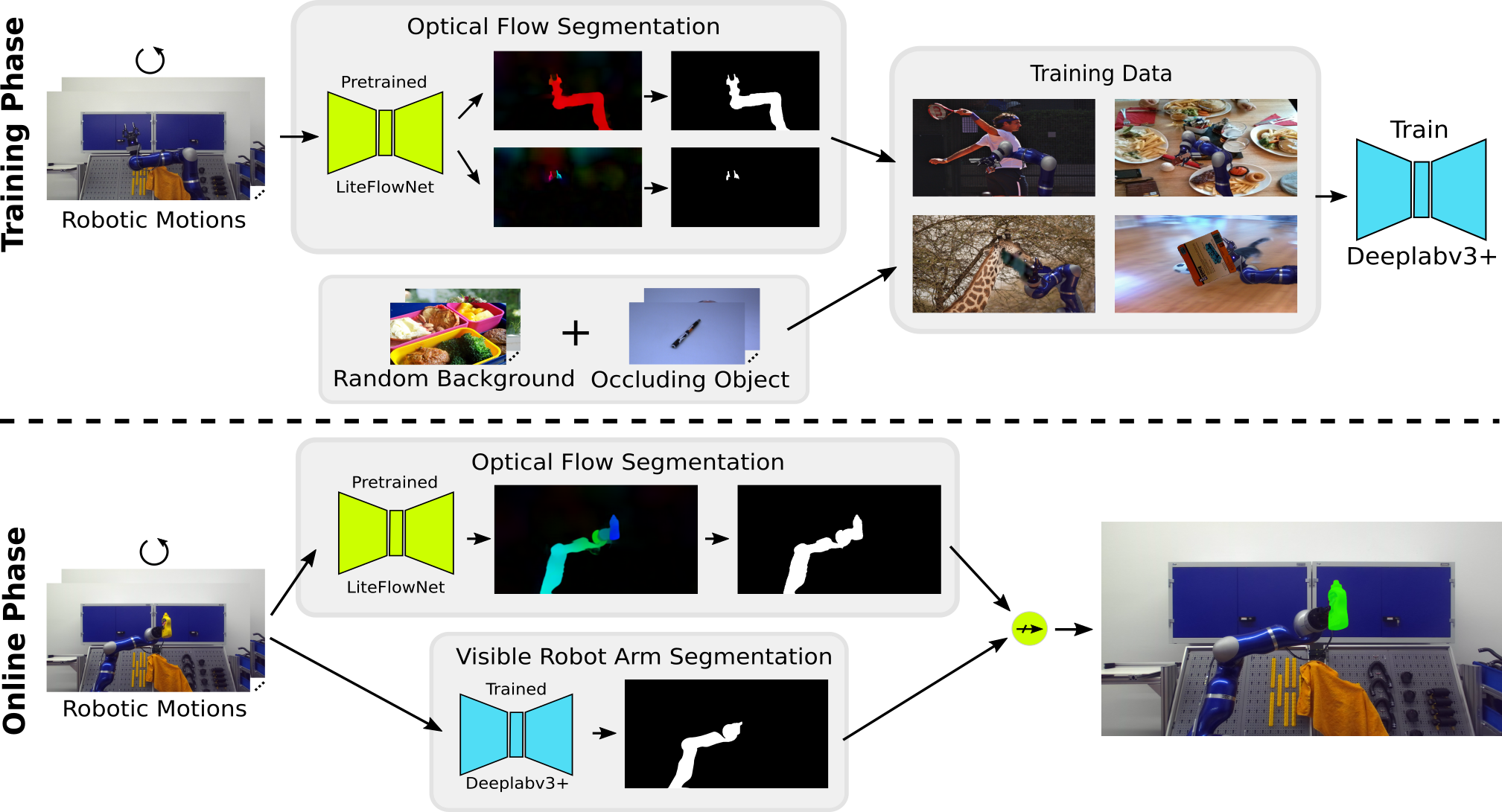}
	\caption{In a training phase we predict optical flow for robot arm motions using a pre-trained LiteFlowNet. By thresholding pixel-wise velocity we generate robot masks that are used to train a DeepLabv3+ to segment the visible robot arm from randomized backgrounds. We emulate the presence of an occluding object at the gripper's position  (top). The trained CNN is then used to separate the robot manipulator from any unknown, grasped object in the joint optical flow (bottom), where~$\nrightarrow$~denotes a nimply gate. The result is an abundant source of labeled data that can be produced online in any static environment.}
	\label{fig:front2}
\end{figure*}

In this section, we describe in detail how a combination of optical flow and learned segmentation can be leveraged to automatically segment both, a robotic arm and a grasped object, from single RGB camera streams only. See Figure \ref{fig:front2} for an overview.

\subsection{Optical Flow Segmentation}
\label{sec:ofs}


We employ a LiteFlowNet \cite{of_liteflownet2018hui} trained  on synthetic data
which is able to efficiently estimate optical flow between two images whilst generalizing well to real environments.
By thresholding the velocity magnitude we segment a moving robot arm either with or without a grasped object from static background. 
Instead of a predefined, hard-coded threshold which would constrain us to a specific velocity or camera resolution we apply Otsu's binarization \cite{otsu1979threshold} on the flow magnitude. 
In order to counteract under-segmentation of the LiteFlowNet we take the joint mask of thresholded forward~\mbox{(\emph{t $\rightarrow$ t+1})} and backward \mbox{(\emph{t $\rightarrow$ t-1})} flow to obtain a segmentation mask.

\subsection{Visible Robot Arm Segmentation}
\label{sec:method_rob_seg}
Still, optical flow alone is
unsuitable to distinguish a robot arm from an unknown, grasped object.
Therefore, we propose to learn how to segment the robot arm from single color images. 
Diverse training data for this task can be automatically generated from sequences of a moving robot arm without any object in its gripper by following the procedure explained in Section \ref{sec:ofs}.
To prevent over-fitting on the background, we paste the extracted robot arm masks at random scale and horizontal translation on images of the MS COCO dataset \cite{dataset_mscoco2014lin}. 
Unlike \cite{florence2019self} we do not incorporate backgrounds from the test environment into our training data.
Furthermore, to emulate the presence of a grasped object we paste one random object crop (different from our test objects) per training sample at the gripper's position.  
We derive these spots by opening or closing the gripper while the robot remains at a certain joint position and obtain a corresponding mask again by thresholding the flow magnitude - a procedure which is not necessary during inference.
Whenever this gripper mask consists of two separated parts, we can further realistically emulate the grasp by placing the object above the smaller part (most likely further away from the camera, thus occluded) and overlay the item with the larger one (potentially closer to the camera, thus occluding the object).
In all other cases we sample a random point on the gripper mask which denotes the object center. The occluding objects are not learned explicitly but belong to the background class. 
Our validation set consists of a disjoint set of robot poses and occluding objects.
As we want to verify the generalization capability of our approach onto novel environments, we do not replace the background for these images.
We generate 45,000 training and 5,000 validation images and train a DeepLabv3+ \cite{chen2018encoder} on semantic segmentation. Afterwards, we are able to segment the visible robot arm regardless of the object in the gripper.
Information on the occluder objects and additional training details are given in Section \ref{app:training}.



\subsection{Grasped Object Segmentation}
\label{sec:method_obj_seg}
During run time, we use the same flow network to derive a joint robot-object mask. Concurrently, we predict the visible robot arm with the trained Deeplabv3+.
Finally, parts in motion but not detected as robot arm are denoted as the unknown object.
We further refine object masks to encounter false-positive and false-negative predictions of both networks.
First, we delete all parts in contact with the image border. We expect the object to be fully visible in the camera frame, and we hypothesize that our trained network successfully segments the area around the gripper. Hence, the object mask should be detached from all irrelevant mask fractions identified by optical flow thresholding.
Second, we once more utilize the gripper masks which were derived by opening and closing the gripper at stationary robot arm positions during the visible robot arm segmentation. These spots are now being used as estimation of the object's position:
We always keep the part of the mask that is closest to this location, but reject every component that is further away than 100~pixels.
Finally, we delete mask parts whose area is below 2500 pixels to remove small noisy artifacts.

\subsection{Trajectory Generation}
\label{sec:setup_traj}

The main goal of our method is to automatically generate data for training object segmentation networks and for performing 3D object reconstruction. For both tasks it is important to obtain as diverse object views as possible. A trajectory obtained e.g. by kinesthetic teaching could instead create object views biased towards specific poses.

Even though we do not assume that the object-camera transformation is known, we can still produce nearly equally distributed end-effector rotations using Fibonacci sampling \cite{gonzales2009measurement} \mbox{with $N=301$ 2D points} on a sphere (mathematical details are explained in Section \ref{app:fibonacci}). 
%
From these points 
we can construct corresponding rotation matrices $R_i$ via up and forward vectors. The robot would heavily occlude the object in some of the poses where the gripper points away from the camera. Therefore, we split the view sphere in half and mirror the gripper rotations pointing away from the camera to the side that points approximately towards the camera. To obtain views from all sides, the motion is repeated with the gripper rotated 180 degrees. 


Additionally, some rotational end-effector movements do not create sufficient motion to reliably detect the object mask by means of optical flow. 
Yet, motion between two consecutive frames is of severe importance for successful segmentation. To alleviate this dependency one could propagate optical flow over multiple frames (e.g. \cite{mai2017refinement}). 
However, this can lead to drift effects and propagation of erroneous regions, which potentially worsens an initial prediction.
Therefore, we propose to continuously follow Cartesian points on an ellipse-shaped trajectory that is approximately parallel to the image plane. We repeatedly loop through $n_e=20$ equidistant points on the ellipse and use them as end-effector translations. Finally, we assign end-effector rotation matrices $R_i$ by spiraling through the points on the sphere from the Fibonacci sampling.

\section{Setup and Dataset Creation}

The main setup consists of a KUKA LBR4+ robot arm 
on a linear axis with a Robotiq 2F-85 two-finger gripper. 
Images are recorded with a ZED stereo sensor.

\subsection{Robotic Arm Recordings under Weak Camera Calibration}
\label{sec:rad}

A major advantage of our method is the independence of a precise robot-camera calibration.
Yet, 
we expect the robot to move inside the camera frame on an ellipse-shaped trajectory which should be approximately parallel to the image plane (Section \ref{sec:setup_traj}). Furthermore, all parts of the object should be visible inside the frame's borders to not be discarded during post-processing (Section~\ref{sec:method_obj_seg}). 
Nevertheless, these constraints can be satisfied quite effortless with approximate knowledge about the robot's motion in the world coordinate frame, and by placing the camera loosely pointing towards the robot.
We use the trajectory as described in Section \ref{sec:setup_traj} to record images of the robot at 301 poses. At every configuration $P_i$ we shift the linear axis with a small offset to ensure that all parts of the robot jointly move at all times. Additionally, at every such pose we solely move the gripper jaws to different positions to determine spots where to paste occluding objects, as explained in Section \ref{sec:method_rob_seg}.

\subsection{Segmenting YCB-Video Objects}

To demonstrate the effectiveness of our suggested method we record 15 different objects of the YCB-Video Dataset \cite{methods_ycbvideodataset2017xiang} while being grasped by a robotic gripper. 
Grasping of unknown objects is a challenging task on its own and often requires priors in the form of object poses \cite{methods_autgrasp2014shi}. 
As this is not the focus of our work
we directly hand over the respective objects to the robot which could also be a realistic application scenario. Nevertheless, to further automate our approach we carry out a grasping study on these objects and refer the reader to Section \ref{app:grasping} of the supplementary material for results.
%
Note that we use the same trajectory as in Section \ref{sec:rad} for a direct comparison of our method to change detection algorithms.
\subsection{Baselines: Change Detection and Object-Agnostic Segmentation}
\label{sec:method_cd}

As a baseline measure we explore pixel-wise change detection \cite{cd_definition1989singh, cd_survey2005radke} in RGB space (CD\textsubscript{RGB}) and between optical flow masks (CD\textsubscript{OF}), and direct object-agnostic segmentation (Obj-Agn). 
For change detection we record images of the robot arm both with and without an object at the same position and select pixels based on the absolute difference between them. In the RGB case we empirically set $\tau = \frac{p}{25}$ with $p$ denoting the pixel range; for optical flow masks this reduces to a binary selection.
%
%
Regarding direct object-agnostic segmentation we train another DeepLabv3+ on semantic segmentation (similar to the one trained on robot arm segmentation), and additionally predict the object in the gripper as separate semantic class. During testing we can directly infer the binary mask of the unknown object from one single RGB frame.
For fair comparison we apply the same post-processing on every method except for change detection in RGB space, where we perform morphological opening and closing.

\section{Evaluation of Automatically Annotated Object Masks}

\label{sec:experiments}
We proceed to evaluate our automatic annotation method. For all segmentation results we report the standard \textit{mean Intersection over Union} (mIoU). Therefore, we manually annotate 15\% of all recorded YCB object images with ground truth segmentation masks.
The object poses of this subset are identical across all items.

\begin{figure*}[!t]
	\centering
	\subfloat[][RGB]{\includegraphics[width=0.14\textwidth]{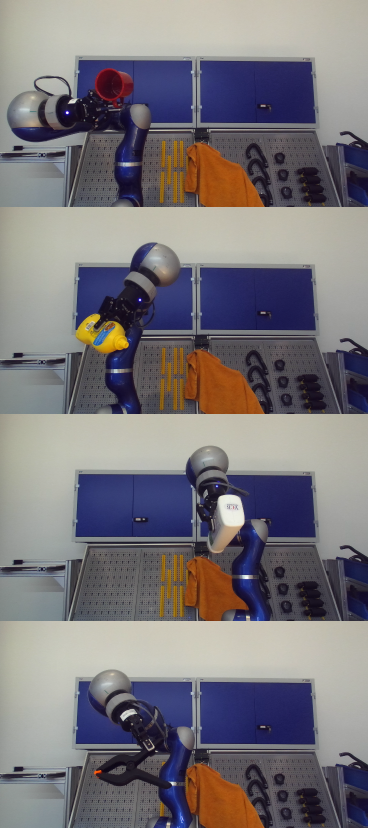}}
	\subfloat[][Fwd. Flow]{\includegraphics[width=0.14\textwidth]{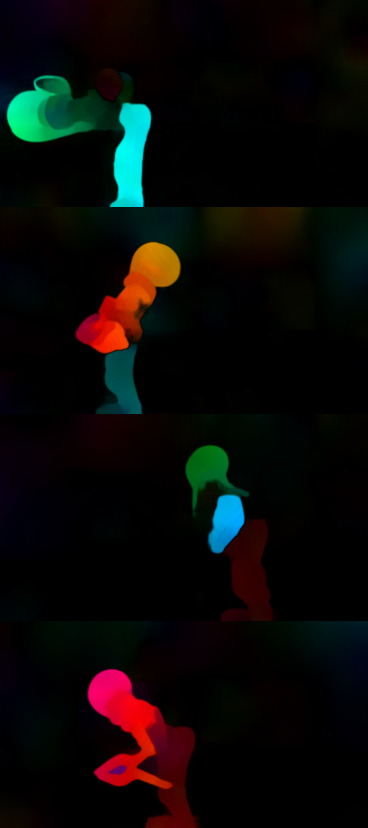}}
	\subfloat[][Flow Mask]{\includegraphics[width=0.14\textwidth]{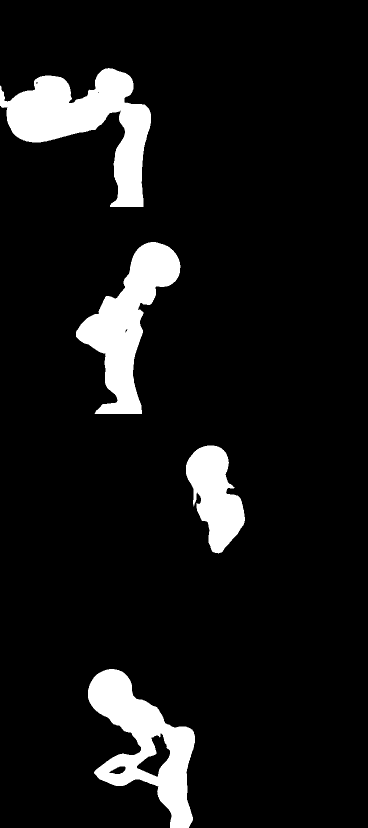}}
	\subfloat[][CD\textsubscript{RGB}]{\includegraphics[width=0.14\textwidth]{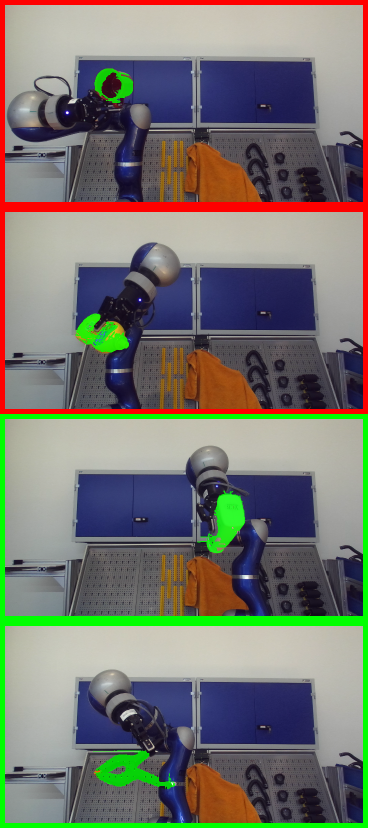}}	
	\subfloat[][CD\textsubscript{OF}]{\includegraphics[width=0.14\textwidth]{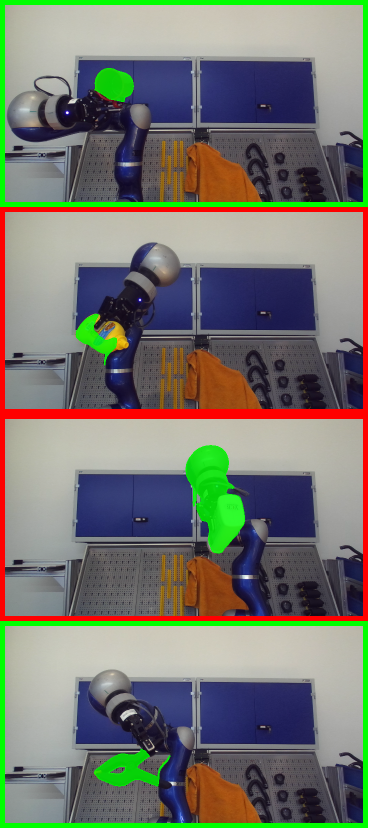}}
	\subfloat[][Obj-Agn]{\includegraphics[width=0.14\textwidth]{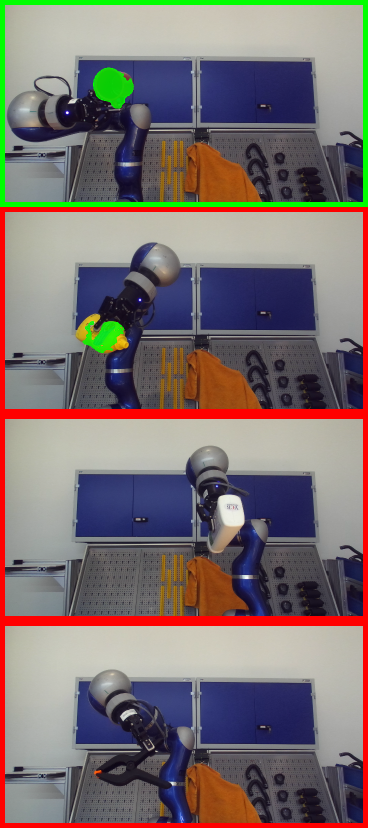}}
	\subfloat[][Ours]{\includegraphics[width=0.14\textwidth]{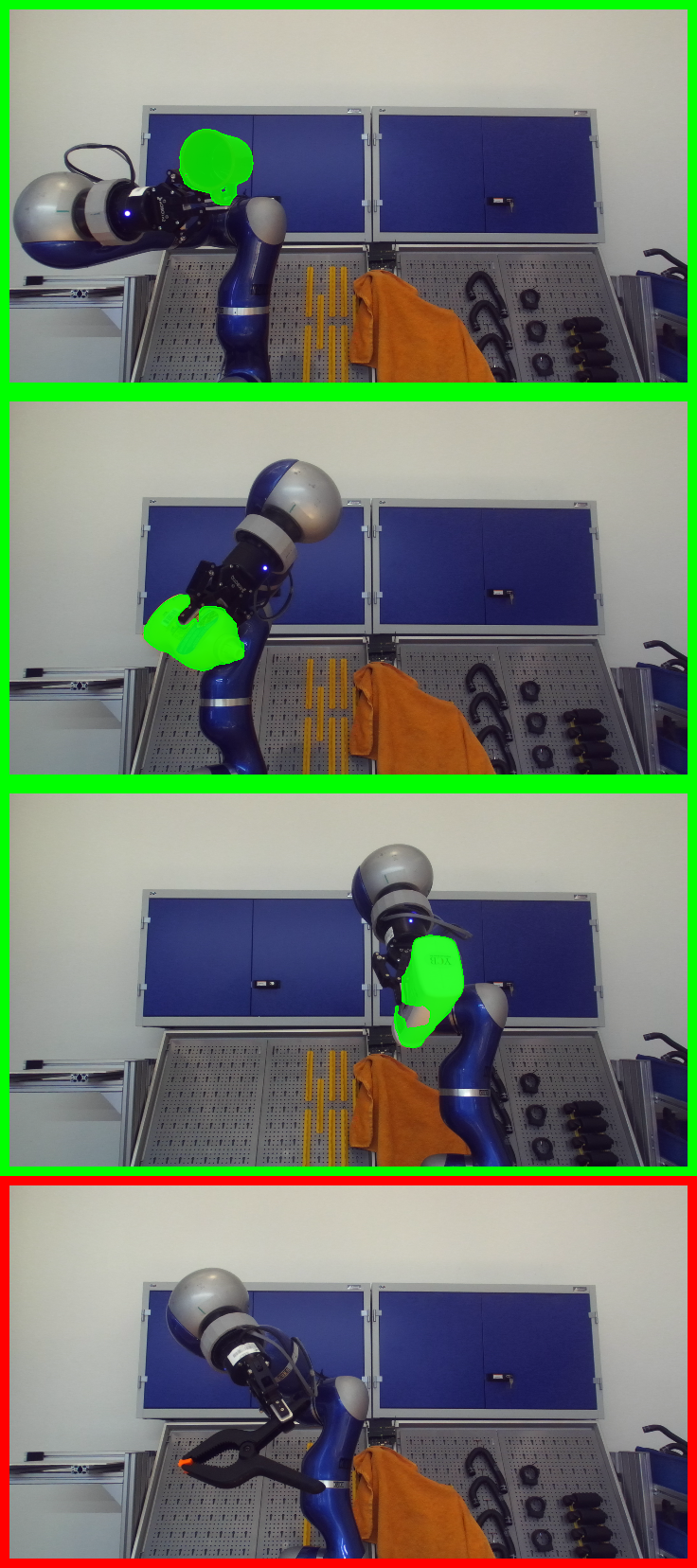}}
	
	\caption[Qualitative results]{Qualitative results of our method and the three baselines (best viewed magnified and in color). Red borders denote poor performance, while green ones indicate satisfactory segmentation masks. Top row: CD\textsubscript{RGB} tends to perform poorly whenever object and background are of similar color. Note that while CD\textsubscript{OF} and Obj-Agn deliver decent results despite small cut-off parts, our method even successfully segments the handle (best viewed zoomed-in). Second row: CD\textsubscript{OF} can only segment moving parts as object which are directly located in front of the background. Third row: CD\textsubscript{OF} over-segmentation due to faulty LiteFlowNet predictions, which our method corrects.
		Fourth row: A failure case of our method where parts of the object are identified as robot due to the similar color and texture of clamp and gripper. Our applied post-processing deletes the remaining small blobs and results in an empty mask. Obj-Agn fails to segment the last two objects completely.}
	
	\label{fig:qualitative_results}
\end{figure*}



\subsection{Comparison to Baselines}
\label{sec:exp_baseline}

Table~\ref{tab:object_masks} reports the mIoU of our proposed method compared to three baselines. We outperform these by a considerable margin across all objects except one, which is best explained in Figure \ref{fig:qualitative_results}. 
\begin{table*}[!ht]
	\centering
	\caption[Quantitative results]{Quantitative comparison of our method compared to three baseline approaches and \cite{florence2019self}; for a better overview we first list the joint objects. The two right-most columns show results on semantic segmentation and detection in a table top scene, where we use our automatically generated object masks as training data. Note that we compute the IoU for detection for a comparison to the semantic segmentation results - this is possible since no instance appears more than once per image. Numbers in \textbf{bold} denote the best results on object-in-gripper segmentation.}
	\vspace*{1mm}
	\resizebox{0.9\textwidth}{!}{
		\input{quant_results_full}}
	\label{tab:object_masks}
\end{table*}
The performance gain can be traced back to the fact that optical flow and robot segmentation networks are trained to extract features that are robust against noise, similar colors and non-rigid parts.
We also circumvent the hard, potentially ill-defined task of learning how to segment unknown objects - here,  
an object is simply defined as a physically connected entity in motion.
For further results on deformable objects like towels and detailed evaluation of our post-processing we refer the reader to Section \ref{app:deformable} of the supplementary material.

\subsection{Comparison to Literature}
To the best of our knowledge, the most similar approach in literature stems from Florence et al. \cite{florence2019self}. Yet, they require robot-camera calibration as well as depth information and obtain their foreground mask with GrabCut, while the distinction between robot and object is done in a similar fashion with a CNN. 
We compare our method on a joint subset of YCB objects (see Table \ref{tab:object_masks}).
Although we refrain from camera calibration and solely utilize a single RGB image stream, we are able to surpass their respective baseline.
\subsection{Object Segmentation from Scenes}

\begin{figure}
	\begin{minipage}[t]{0.42\textwidth}
		\vspace{0pt}
		\includegraphics[width=\linewidth]{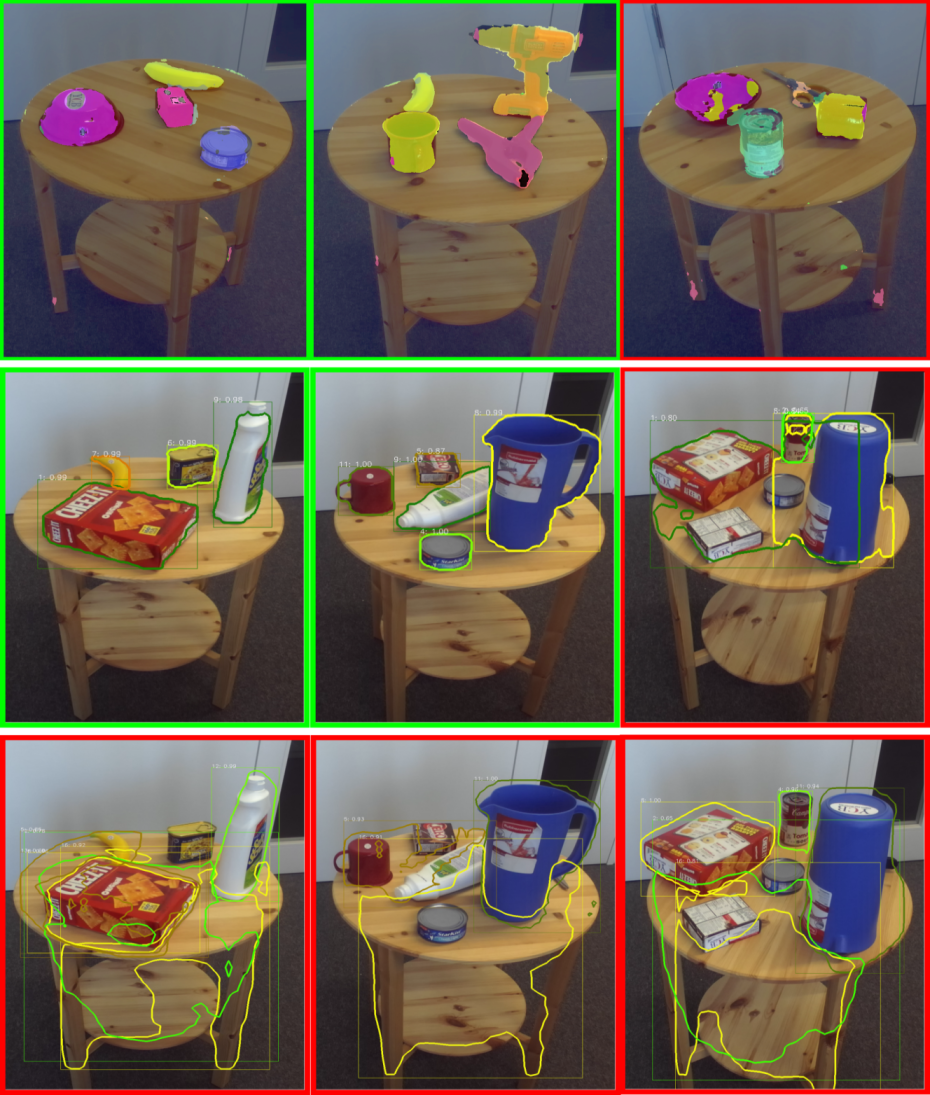}
		\captionof{figure}{Semantic / instance segmentation results. Top row: DeepLabv3+ trained on our grasped objects on semantic segmentation; 
			Middle row: MaskRCNN trained on our grasped objects;
			Bottom row: MaskRCNN trained on the YCB Video dataset \cite{methods_ycbvideodataset2017xiang}.
			Networks trained on our acquired data usually generalize well to table top scenes with few failure cases due to thin parts or over-segmentation (first two rows). 
			Networks trained on the original annotated YCB data, which consists of table top scenes recorded with a different sensor, fail to generalize due to the domain gap (bottom row).}
		\label{fig:semseg}
	\end{minipage}%
	\hfill
	\begin{minipage}[t]{0.54\textwidth}
		\vspace{0pt}\raggedright
		\includegraphics[width=\linewidth]{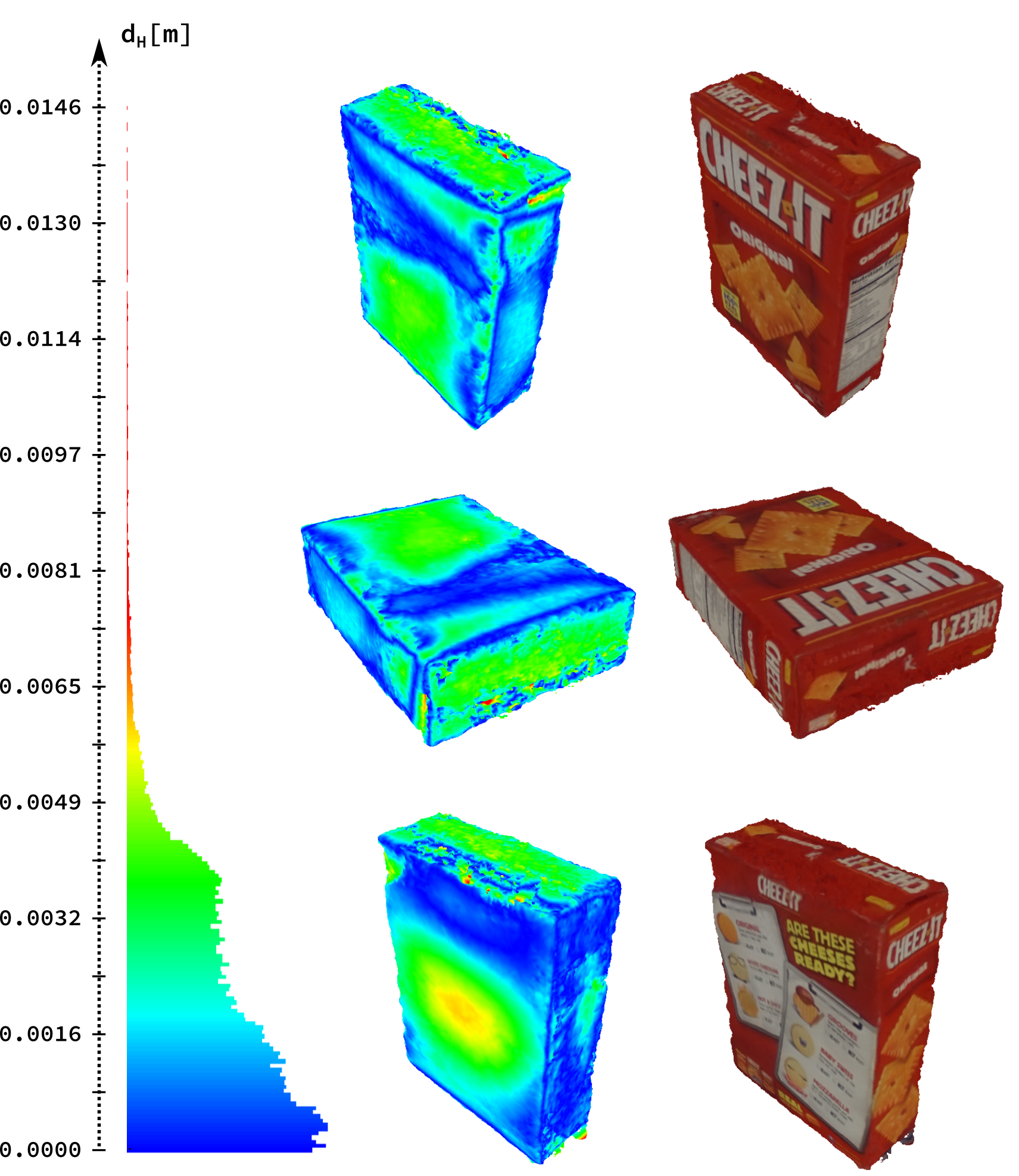}
		\captionof{figure}{Photogrammetric 3D Reconstruction of \emph{003\_cracker\_box} directly from the robotic gripper; Left: Histogram of Euclidean distances $d_H$ from nearest vertices between ground truth mesh (YCB Video dataset \cite{methods_ycbvideodataset2017xiang}) and our aligned 3D reconstruction. The mean distance between vertices is $\mu(d) = 0.2 cm$; Middle: Quality Mapping; Right: Texture Mapping; Note that any potential gripper parts were successfully removed from the object.}
		\label{fig:reconst}
	\end{minipage}
\end{figure}

We use the generated object masks to train DeepLabv3+ on semantic segmentation from RGB scenes. We test our model on 100 images where we place the respective objects at random position and orientation on a table. Results are depicted in Figure \ref{fig:semseg} and in the second from right column in Table~\ref{tab:object_masks}. 
Even though the test set was recorded in a different environment than the training data, the mIoU values in the test scene are approaching those of the generated in-gripper training data. The \textit{pitcher} and \textit{drill} even receive higher mIoU scores than when segmented in the gripper. We ascribe this to the ability of neural networks to average out annotation errors. 
However, although the \textit{mug} and \textit{bowl} stood out during the automatic mask generation, the network seems to have difficulties differentiating between them since they have very similar texture. We believe that this is because of incomplete masks which provide unreliable information on the object's shape.

We also train MaskRCNN~\cite{he2017mask} on instance segmentation using the same automatically annotated training data (right-most column of Table~\ref{tab:object_masks}) and qualitatively compare results to a MaskRCNN trained on the YCB Video dataset in Figure \ref{fig:semseg}. It shows that our grasped object data generalizes to novel settings while training on YCB Video does not generalize to a novel sensor. 

CNNs typically require many different views of an object to derive a good representation that generalizes well to novel scenes. To this end, we investigate the performance loss onto the same test scene given less than the initial $n=301$ almost equidistant views. While merely two views of an object provide sufficient information to achieve 60\% of the final mIoU with 301 views, the results underline that view diversity indeed increases segmentation performance. For more results please see Section \ref{app:views} of the supplementary material.



%
%

\subsection{Photogrammetric 3D Object Reconstruction}

Textured 3D meshes contain rich information about objects. In robotics they can be utilized for grasp planning, object pose estimation or tracking and learning in simulation. A robot that can perform 3D object reconstruction without human intervention therefore strongly increases autonomy.

Photogrammetric 3D object reconstruction is usually performed by taking photos of a static object from various camera positions, finding correspondences, computing their 3D coordinates and thereof building a mesh. Disadvantages of this method are that the bottom of the object is usually not captured and that the environment and object should be static at all times. Since our presented methods yield accurate segmentation masks, we can reverse the problem by actively recording views of a rotating object using a static camera. One challenge is that the lighting on the object is dynamic during the recording process. Nonetheless, our masked object views can be successfully fused without manual intervention or tuning using a commercial reconstruction software \cite{agisoft2018agisoft}. Figure~\ref{fig:reconst} shows the resulting textured mesh of the YCB Cracker Box and the Euclidean distances of the nearest vertices on the aligned ground truth model. Please note that pure monocular photogrammetric 3D object reconstruction can also fail for some symmetric and texture-less objects. In these cases, information from depth cameras has to be incorporated.

%
%

%


\section{Conclusion}
\label{sec:conclusion}
We have presented a framework to automatically generate unseen object and robot arm segmentation masks through robotic interactions. Our optical flow based approach requires little prior knowledge and poses few constraints on the setup. Particularly, the camera does not need to be calibrated against the robot and thus can be freely moved. 
The usability of the generated labels has been demonstrated in several downstream tasks like semantic and instance segmentation from scenes and 3D object reconstruction. The latter is a step forward to close the real2sim2real cycle.
Our work is motivated by both the way humans learn through object interactions and practical limitations of supervised learning algorithms that stem from the lack of annotated data. We conclude that observing self-generated motions can enable robots to semi-autonomously learn about newly encountered objects, which is a prerequisite for self-improving systems.

\clearpage
\acknowledgments{We thank the reviewers for their useful comments. This work was partially supported by the \mbox{DLR-internal~project} "Factory of the Future".}


\bibliography{refs_short}  

\cleardoublepage  

\appendix



\section{Additional Results}
\subsection{Deformable Objects}
\label{app:deformable}
\begin{figure}[h]
	\begin{minipage}[t]{0.385 \textwidth}
		\vspace{0pt}
		\includegraphics[width=1\linewidth]{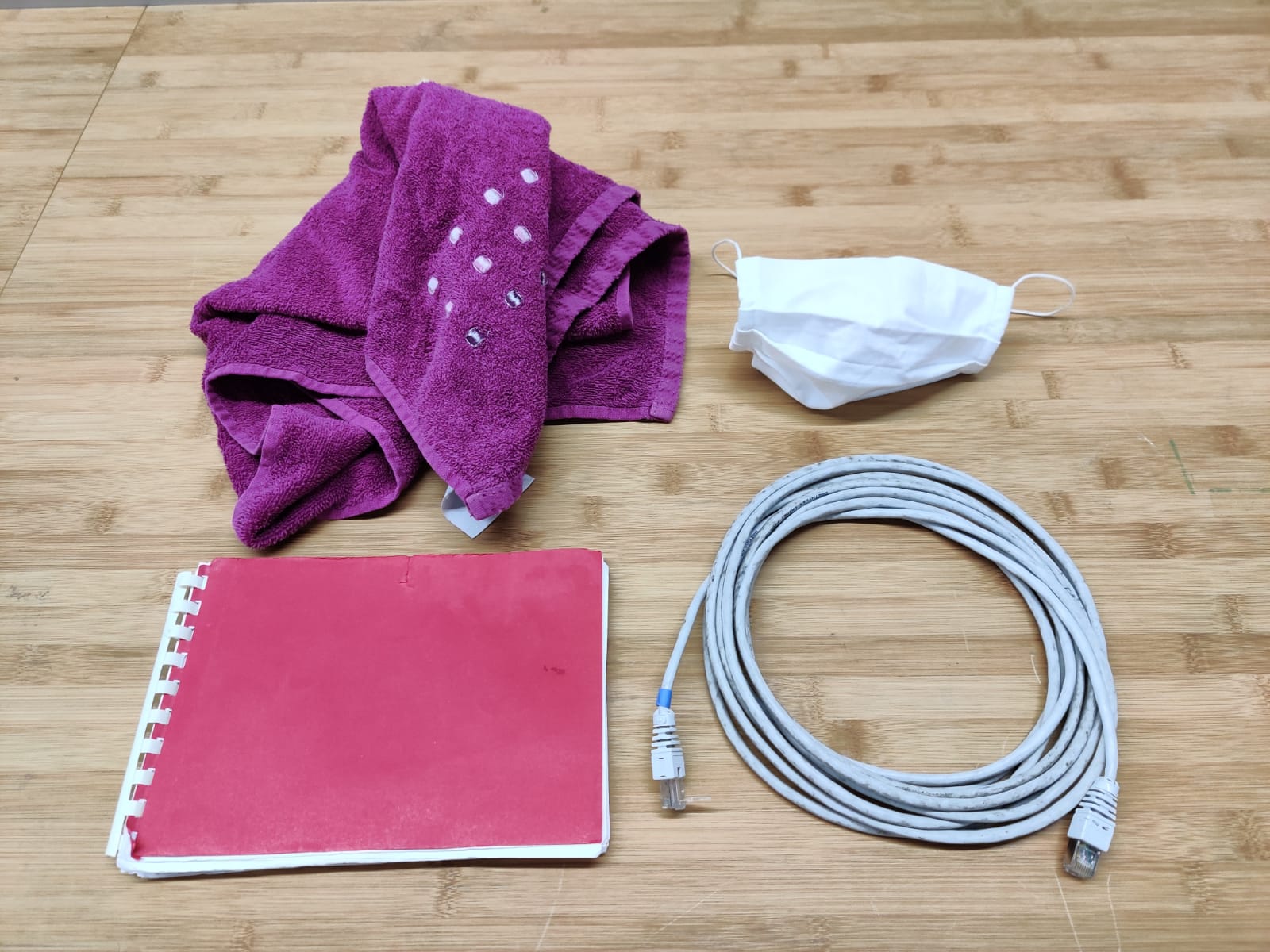}
		\captionof{figure}{Four additional, non-rigid objects. From left to right, top to bottom: towel, mask, notebook, cable.}
		\label{fig:app_deformable}
	\end{minipage}%
	\hfill
	\begin{minipage}[t]{0.57\textwidth}
		\vspace{0pt}\raggedright
		\includegraphics[width=1\linewidth]{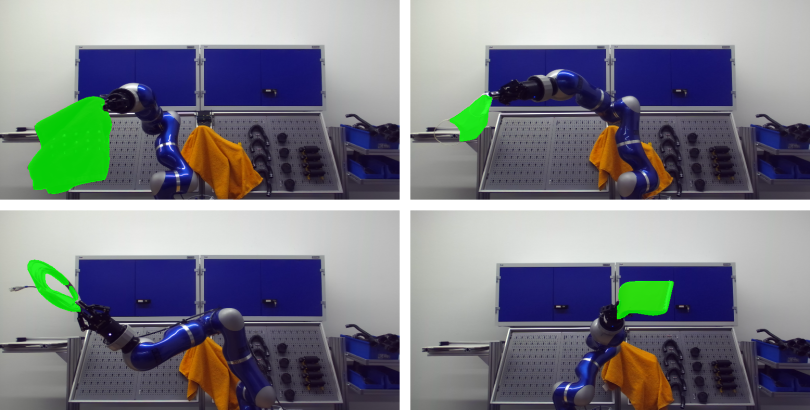}
		\captionof{figure}{Exemplary quantitative results on the four deformable objects. From left to right, top to bottom: towel, mask, cable, notebook.}
		\label{fig:app_deformable_results}
	\end{minipage}
\end{figure}

Our proposed method does not require strict rigid objects. To demonstrate this we record four deformable objects (Figure \ref{fig:app_deformable}) on half of the points in the proposed trajectory (i.e., one half side of the sphere in Figure \ref{fig:fibonacci}). For results see Table \ref{tab:app_deformable} and Figure \ref{fig:app_deformable_results}.

\begin{table}[h]
	\centering
	
	\caption[Deformable]{Additional quantitative results on deformable objects. All values are in \%.}
	
	\begin{tabular}{lccc}
		\toprule
		Deformable Object & mIoU &Precision&Recall\\
		\midrule
		cable & 50.95&63.44&72.13\\
		mask & 69.79 & 72.10 & 95.63\\
		notebook & 77.24&93.41&81.69\\
		towel & 81.81&93.19&87.02\\
		\midrule
		Average & 69.95&80.53&84.12\\
		\bottomrule
	\end{tabular}
	
	\label{tab:app_deformable}
\end{table}

%
%
%

\subsection{Number of Views for Segmentation Performance}
\label{app:views}
\begin{figure}[!h]
	\centering
	\captionsetup{width=\linewidth}
	\includegraphics[width=0.6\linewidth]{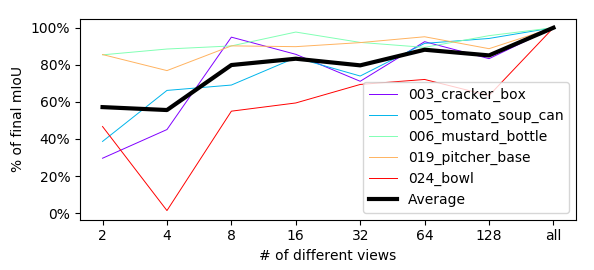}
	\caption[Qualitative results]{Number of different views in relation to the relative final mIoU with all object views for five different objects. With only two viewpoints the network achieves on average 60\% of the final mIoU (301 poses), and there is a notable correlation between performance and number of views.}
	\label{fig:howmany_views}
\end{figure}

It is expected that many distinct views are beneficial for the generalization ability of neural networks. We explore this by training a network with our automatically segmented objects on semantic segmentation in a table-top setting, and increment the amount of object views across different trainings. Figure \ref{fig:howmany_views} shows the respective results.

\section{Grasping of Unknown Objects}

\label{app:grasping}
To further automate our segmentation pipeline we evaluate grasping and re-grasping after placing success on the YCB objects in Table \ref{tab:grasp}. The re-grasping at $180^o$ degree gripper rotation is necessary as we want to record the backside of the object. Here, solely the object's position on the table is known to the robot. Only the \textit{bowl} cannot be grasped at a static position from top-down. The \textit{scissors} is unstable and slightly offset when placed back onto the table so that the re-grasp at $180^o$ gripper rotation can fail. Apart from these issues, the results confirm that our strategy is successful for almost all objects and depicts a practical way to minimize human intervention.

\begin{table}[h!]
	\centering
	\caption[Top-Down grasp success]{Top-Down grasp successes for YCB objects. \textit{Re-graspable} refers to placing the object and re-grasping it at $180^o$ rotated gripper position.}
	\vspace*{5mm}
	\begin{tabular}{lcc}
		\toprule
		YCB Object           & graspable &  re-graspable \\
		\midrule                                                  
		003\_cracker\_box       &  \textcolor{Green}{\cmark}  &        \textcolor{Green}{\cmark}                       \\
		005\_tomato\_soup\_can    &  \textcolor{Green}{\cmark}  &        \textcolor{Green}{\cmark}                       \\
		006\_mustard\_bottle        &  \textcolor{Green}{\cmark}  &        \textcolor{Green}{\cmark}                       \\
		007\_tuna\_fish\_can       &  \textcolor{Green}{\cmark}  &        \textcolor{Green}{\cmark}         \\              
		008\_pudding\_box         &  \textcolor{Green}{\cmark}  &        \textcolor{Green}{\cmark}                       \\
		010\_potted\_meat\_can       &  \textcolor{Green}{\cmark}  &        \textcolor{Green}{\cmark}                       \\
		011\_banana         &  \textcolor{Green}{\cmark}  &        \textcolor{Green}{\cmark}                       \\
		019\_pitcher\_base        &  \textcolor{Green}{\cmark}  &        \textcolor{Green}{\cmark}                       \\
		021\_bleach\_cleanser     &  \textcolor{Green}{\cmark}  &        \textcolor{Green}{\cmark}                       \\
		024\_bowl           &  \textcolor{red}{\xmark}    &        \textcolor{red}{\xmark}                       \\
		025\_cup            &  \textcolor{Green}{\cmark}  &        \textcolor{Green}{\cmark}                       \\
		035\_power\_drill          &  \textcolor{Green}{\cmark}  &        \textcolor{Green}{\cmark}                       \\
		037\_scissors        &  \textcolor{Green}{\cmark}  &        \textcolor{red}{\xmark}                       \\
		052\_extra\_large\_clamp    &  \textcolor{Green}{\cmark}  &        \textcolor{Green}{\cmark}                       \\
		061\_foam\_brick     &  \textcolor{Green}{\cmark}  &        \textcolor{Green}{\cmark}                       \\	
		\midrule
		Total & $14/15$ & $13/15$ \\
		\bottomrule
	\end{tabular}
	\label{tab:grasp}
\end{table}

\section{Fibonacci Sampling}

\label{app:fibonacci}

\begin{figure}[!h]
	\centering
	\captionsetup{width=\linewidth}
	\includegraphics[width=0.8\linewidth]{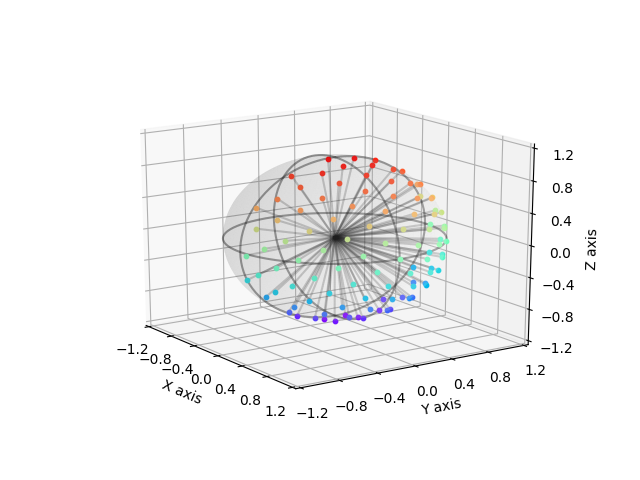}
	\caption{Visualization of Fibonacci sampling of almost equidistant points on a sphere. We only consider the end-effector rotations pointing approximately towards the camera and then regrasp the object for the back side of the sphere.}
	\label{fig:fibonacci}
\end{figure}

First we create an evenly distributed lattice using $N=301$ 2D points with coordinates

\begin{equation}
(x_i,y_i) =   \left(  \frac{i+1/2}{N},  \frac{i}{\phi} \right)  \quad \textrm{for }\; 0 \leq i \leq N-1 \tag{2}
\end{equation}
where $\phi  = \frac{1+\sqrt{5}}{2}$ is the golden ratio. Then, we perform an area preserving mapping onto a cylinder and subsequently onto a sphere by
\begin{equation}
(x,y) \rightarrow (\theta, \phi) : \quad  \left( \cos^{-1}(2x-1) – \pi/2,  2\pi y \right)
\end{equation}
\begin{equation}
(\theta,\phi) \rightarrow (x,y,z) : \quad \left (\cos\theta \cos\phi, \cos \theta \sin \phi, \sin \theta \right)
\end{equation}

to obtain the respective coordinates on the sphere. We construct corresponding rotation matrices $R_i$ via up and forward vectors.
A visualization of a reduced set of almost equidistant points on a sphere is depicted in Figure \ref{fig:fibonacci}.

\subsection{Ablation Study of our Post-Processing}
\label{app:abl_pp}
In Table \ref{tab:ablation} we investigate the influence of different design choices for our method regarding data generation / training setting and post-processing on the YCB objects. 

\begin{table}[h!]
	\centering
	
	\caption[Ablation study]{Ablation study on different data augmentation and training settings (\textit{D/T}) and post-processing (\textit{PP}) across different flow masks (from which we subtract the robot prediction to derive an object mask). In every consecutive row we drop the respective step and all previous ones. Altogether we are able to boost object-in-gripper segmentation from an initial object mIoU of around 51\% by about 25 percent points. Numbers in \textbf{bold} denote the best results.}
	\input{ablation}
	\label{tab:ablation}
\end{table}

The main success factors for our final metrics are the pasting of a distractor object, the maximum gripper distance and the minimum mask size. 
There is, albeit small, a notable difference between forward flow and forward/backward union, and at its best, forward flow beats cumulative masks by a small margin. However, this is merely a result of discarding information in the last post-processing step and we therefore used the cumulated masks. 
Flow intersection masks between forward and backward flow, on the other hand, might eliminate noise but heavily rely on motion at the same spatial positions between consecutive frames. Even with perfect continuous movement this can not be guaranteed due to potential mis-predictions of the LiteFlowNet. 

\subsection{mIoU, Precision and Recall between Vanilla and Post-Processed Results}

Table \ref{tab:ipr_pp} lists the gain in \% for different metrics for three baselines and our proposed approach.

\begin{table}[h!]
	\centering
	
	\caption[Quantitative results ipr pp]{Quantitative comparison between vanilla results and applied post-processing. The metrics denote the respective average across all object classes. Especially noteworthy are the small differences between our vanilla and CD\textsubscript{OF} post-processed results. All values are in \%. Numbers in \textbf{bold} denote the best results.}
	\vspace*{5mm}
	\input{ipr_pp}
	\label{tab:ipr_pp}
\end{table}

\section{Training Details}
\label{app:training}

\begin{figure}[!h]
	\centering
	\captionsetup{width=\linewidth}
	\includegraphics[width=0.6\linewidth]{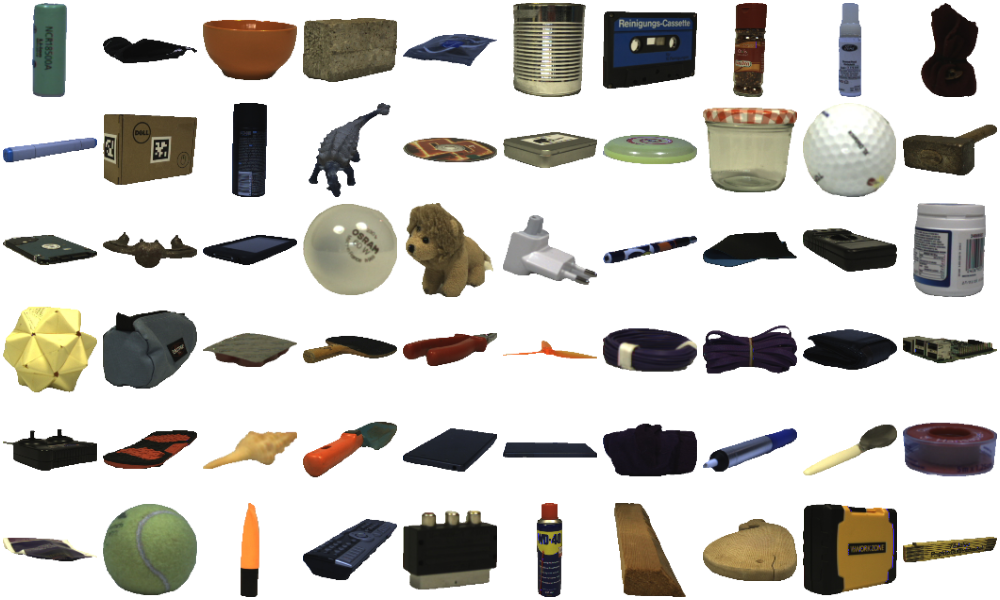}
	\caption{Occluding objects used in our training data.}
	\label{fig:occl_obj}
\end{figure}

As occluding objects we use objects from the BigBird dataset (\href{http://rll.berkeley.edu/bigbird/}{http://rll.berkeley.edu/bigbird/}) (discarding instances of the YCB Video dataset), and a set of internally recorded objects depicted in Figure \ref{fig:occl_obj}.
During training we apply random color jitter on the occluding objects with focus on blueish augmentation to account for over-segmentation on our blue-biased robot arm. 
To ensure that the model is able to differentiate well between gripper and object, we weight the area around the gripper with a Gaussian shape when calculating the loss. The respective center is again obtained from the gripper spots. We empirically set the width to 50 pixels and the magnitude at the center to a factor of 3.

We use a DeepLabv3+ implementation in PyTorch pre-trained on ImageNet. For all object-in-gripper segmentation tasks we use the default learning rate with a polynomial decay and Adam optimizer. We train with a fixed input size of 414x736 pixels.

For all optical flow predictions we use a LiteFlowNet implementation (\href{https://github.com/sniklaus/pytorch-liteflownet}{Link to GitHub}) pre-trained on Things3D and Chairs. 
LiteFlowNet is our flow predictor of choice due to its superior performance on popular benchmark datasets at time of release and fast flow estimation.
For an exhaustive comparison between conventional, hybrid and network-based optical flow estimators we refer the reader to \cite{of_liteflownet2018hui}, Table 2.

For semantic segmentation we use the same DeepLabv3+ but freeze the encoder. The decoder is trained with a reduced learning rate of 0.0007. Similar to the robot arm segmentation, we paste multiple instances on MS COCO images at random scale, translation and in-plane rotation and adapt the labels accordingly. We allow occlusions up to 60\% of each object instance mask and re-sample the training image if this limit is exceeded. Thereby, we produce 50,000 images. As the objects differ in size, we weight the object classes inversely proportional by the number of pixels to put equal focus on smaller objects, and clip the background weight to the factor of the largest object class. 

We train the MaskRCNN with the standard settings and refer the reader to the original paper \cite{he2017mask} for further details.

\section{Estimated Setup and Inference Time}
\label{app:setup_time}

Regarding the aspect of setup time vs. manual annotations, users spent on average 61 seconds to draw the outline of a single object for the PASCAL VOC 2010 trainval dataset \cite{is_turk_pvoc2011maji}. 
Assuming similar timings this would result in more than 76 hours if we would manually label every image of our recorded YCB objects (301 images / 15 objects).
Due to the fact that our algorithm needs no precise calibration, we expect implementing the setup to only take a few hours (excluding training time).
Then, a human merely needs to place objects at marked positions for top-down grasping, or hand the respective object to the robot, which should take no longer than a minute per object. Afterwards, there is no additional human involvement and the camera can be moved.
We conclude that while there is a qualitative difference between manual and our self-supervised annotations, our method is to be preferred regarding setup time. 

In inference mode, our method runs at approximately 5 fps on a single Nvidia GeForce \mbox{RTX 2080 Ti}.

\section{YCB Objects used in this work}
\label{app:ycb_objects}
\begin{figure}[h]
	\centering
	\captionsetup{width=\linewidth}
	\includegraphics[width=0.8\linewidth]{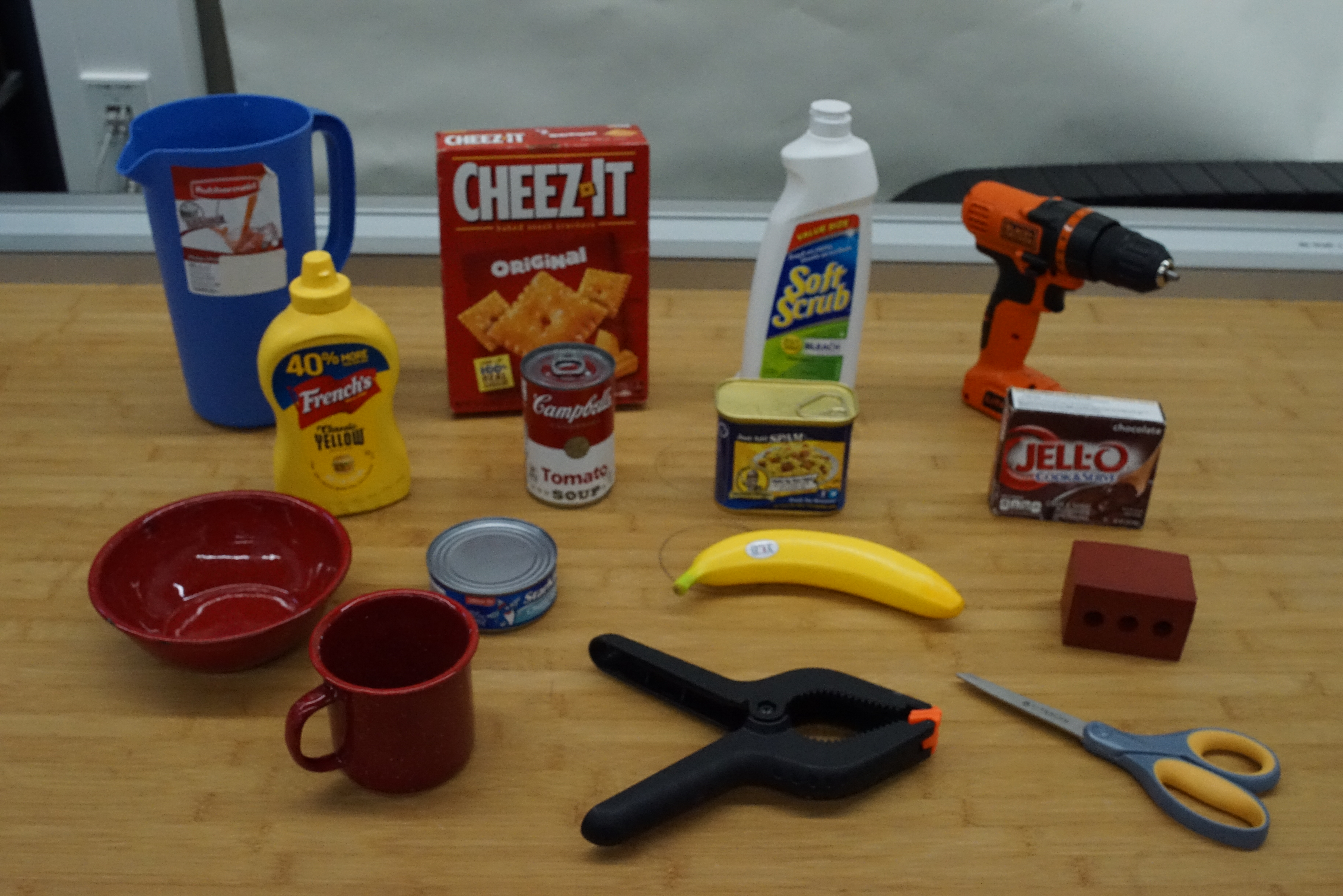}
	\caption{The 15 items from the YCB Video Objects which are used in our experiments.}
	\label{fig:ycb_objects}
\end{figure}

Figure \ref{fig:ycb_objects} shows all 15 YCB Video objects used in this work.

%
%

\end{document}

%% file: quant_results_full.tex
\begin{tabular}{lccccccc}
	\toprule
	\multirow{2}{*}{YCB Object} & \multicolumn{5}{c}{Object-in-Gripper Segmentation} & \multicolumn{2}{c}{\emph{Ours} as training data}\\
	\cmidrule(lr){2-6} \cmidrule(lr){7-8}
	& CD\textsubscript{RGB} & CD\textsubscript{OF} & Obj-Agn & \cite{florence2019self} & Ours & Sem. Seg. & Detection\\
		\cmidrule(lr){1-6} \cmidrule(lr){7-8}
	011\_banana & 33.72 & 47.13 & 00.00 & 54.60 & \textbf{77.23} & 75.18&54.63\\
	019\_pitcher\_base & 42.05 & 74.66 & 48.87 & \textbf{91.90} & 78.56 & 88.75&88.93\\
	024\_bowl & 41.09 & 78.18 & 57.07 & 90.10 & \textbf{91.19} & 53.69&75.99\\
	025\_mug & 39.38 & 77.26 & 84.73 & 84.30 & \textbf{86.36} & 52.68&86.22\\
	035\_power\_drill & 18.16 & 50.77 & 52.32 & \textbf{62.60} & 60.69 & 77.72&79.10\\
	037\_scissors & 06.43 & 10.35 & 00.00 & \textbf{62.70} & 46.01 & 42.35&26.27\\
	061\_foam\_brick & 22.10 & 61.69 & \textbf{87.29} & 57.40 & 80.08 & 74.83&60.56\\
		\cmidrule(lr){1-6} \cmidrule(lr){7-8}
	Average (joint objects) & 28.99 & 57.15 & 42.91 & 71.94 &\textbf{74.30} & 66.46&67.39\\
		\cmidrule(lr){1-6} \cmidrule(lr){7-8}
			\cmidrule(lr){1-6} \cmidrule(lr){7-8}
	003\_cracker\_box & 34.06 & 70.60 & 57.50& - & \textbf{88.84} & 74.29&75.19\\
	005\_tomato\_soup\_can & 27.57 & 38.66 & 76.57 & - & \textbf{81.53} & 62.49&91.70\\
	006\_mustard\_bottle & 32.31 & 66.94 & 55.56 & - & \textbf{86.24} & 79.77&80.92\\
	007\_tuna\_fish\_can & 17.87 & 52.69 & 00.00 & - & \textbf{60.81} & 51.27&42.99\\
	008\_pudding\_box & 46.10 & 64.89 & 76.66 & - & \textbf{80.40} & 73.78&78.64\\
	010\_potted\_meat\_can & 34.81 & 71.07 & 75.46 & - & \textbf{81.78} & 63.87&82.53\\
	021\_bleach\_cleanser & 32.99 & 61.19 & 33.99 & - & \textbf{83.82} & 54.01&83.79\\
	052\_extra\_large\_clamp & 38.98 & 52.37 & 00.00 & - & \textbf{53.48} & 58.68&46.45\\
		\cmidrule(lr){1-6} \cmidrule(lr){7-8}
	Average (total) & 31.18 & 58.56 & 47.07 & - &\textbf{75.80} & 65.56&70.26\\
	\bottomrule
\end{tabular}

%% file: ablation.tex
\begin{tabular}{clccc}
	\toprule
	& &Forward Flow & Intersection & Union\\
	\midrule
	&Ours&\textbf{76.32}&72.15&75.80\\
	\midrule
	\parbox[t]{2mm}{\multirow{3}{*}{\rotatebox[origin=c]{90}{PP}}} &\textcolor{red}{-} Min. mask size &65.54&57.29&\textbf{68.04}\\
	&\textcolor{red}{-} Max. grip. dist.& 60.01 & 53.63 & \textbf{61.50}\\
	&\textcolor{red}{-} Border deletion &57.61 & 53.47 & \textbf{58.29}\\

	\midrule
	\parbox[t]{2mm}{\multirow{2}{*}{\rotatebox[origin=c]{90}{D/T}}} &\textcolor{red}{-} Gripper loss weight &56.24&51.22&\textbf{57.55}\\
	&\textcolor{red}{-} Occluding object&\textbf{51.79}&48.14&51.70\\

	\bottomrule
\end{tabular}

%% file: ipr_pp.tex
\begin{tabular}{lcccccc}
	\toprule
	\multirow{2}{*}{Method} & \multicolumn{3}{c}{Vanilla} & \multicolumn{3}{c}{With Post-Processing}\\
	\cmidrule(lr){2-4} \cmidrule(lr){5-7}
	& mIoU & Precision & Recall & mIoU & Precision & Recall\\
	\midrule
	CD\textsubscript{RGB} & 10.88 & 11.57 & 72.66 & 31.18 & 60.19 & 39.31\\
	CD\textsubscript{OF} & 38.71 & 44.87 & 73.84 & 58.56 & 69.53 & 78.35\\
	Obj-Agn & 34.36 & \textbf{72.22} & 38.24 & 47.07 & 66.94 & 50.27\\
	Ours & \textbf{58.29} & 63.79 & \textbf{85.42} & \textbf{75.80} & \textbf{81.94} & \textbf{90.60}\\
	\bottomrule
\end{tabular}

%% file: paper.bbl
\begin{thebibliography}{36}
\providecommand{\natexlab}[1]{#1}
\providecommand{\url}[1]{\texttt{#1}}
\expandafter\ifx\csname urlstyle\endcsname\relax
  \providecommand{\doi}[1]{doi: #1}\else
  \providecommand{\doi}{doi: \begingroup \urlstyle{rm}\Url}\fi

\bibitem[Hodan et~al.(2019)Hodan, Vineet, Gal, Shalev, Hanzelka, Connell,
  Urbina, Sinha, and Guenter]{hodan2019photorealistic}
T.~Hodan, V.~Vineet, R.~Gal, E.~Shalev, J.~Hanzelka, T.~Connell, P.~Urbina,
  S.~N. Sinha, and B.~Guenter.
\newblock \href{https://arxiv.org/abs/1902.03334}{Photorealistic Image
  Synthesis for Object Instance Detection}.
\newblock \emph{arXiv preprint arXiv:1902.03334}, 2019.

\bibitem[Sundermeyer et~al.(2018)Sundermeyer, Marton, Durner, Brucker, and
  Triebel]{sundermeyer2018implicit}
M.~Sundermeyer, Z.-C. Marton, M.~Durner, M.~Brucker, and R.~Triebel.
\newblock
  \href{http://openaccess.thecvf.com/content\_ECCV\_2018/html/Martin\_Sundermeyer\_Implicit\_3D\_Orientation\_ECCV\_2018\_paper.html}{Implicit
  3d orientation learning for 6d object detection from rgb images}.
\newblock In \emph{Proceedings of the European Conference on Computer Vision
  (ECCV)}, pages 699--715, 2018.

\bibitem[Lobo et~al.(2014)Lobo, Kokkoni, de~Campos, and
  Galloway]{intro_infantslearning2014lobo}
M.~A. Lobo, E.~Kokkoni, A.~C. de~Campos, and J.~C. Galloway.
\newblock \href{https://www.ncbi.nlm.nih.gov/pubmed/24879412}{{N}ot just
  playing around: infants' behaviors with objects reflect ability, constraints,
  and object properties}.
\newblock \emph{Infant Behav Dev}, 37\penalty0 (3):\penalty0 334--351, Aug
  2014.

\bibitem[Nakayama and Loomis(1974)]{of_visualperception1974Nakayama}
K.~Nakayama and J.~M. Loomis.
\newblock \href{https://www.ncbi.nlm.nih.gov/pubmed/4444922}{Optical Velocity
  Patterns, Velocity-Sensitive Neurons, and Space Perception: A Hypothesis}.
\newblock \emph{Perception}, 3\penalty0 (1):\penalty0 63--80, 1974.

\bibitem[Hui et~al.(2018)Hui, Tang, and Loy]{of_liteflownet2018hui}
T.-W. Hui, X.~Tang, and C.~C. Loy.
\newblock \href{https://arxiv.org/abs/1805.07036}{LiteFlowNet: A Lightweight
  Convolutional Neural Network for Optical Flow Estimation}.
\newblock In \emph{Proc. {IEEE} Conf. on Computer Vision and Pattern
  Recognition (CVPR)}, pages 8981--8989, June 2018.

\bibitem[{Bohg} et~al.(2017){Bohg}, {Hausman}, {Sankaran}, {Brock}, {Kragic},
  {Schaal}, and {Sukhatme}]{perception_ipsurvey2017bohg}
J.~{Bohg}, K.~{Hausman}, B.~{Sankaran}, O.~{Brock}, D.~{Kragic}, S.~{Schaal},
  and G.~S. {Sukhatme}.
\newblock \href{https://ieeexplore.ieee.org/document/8007233}{Interactive
  Perception: Leveraging Action in Perception and Perception in Action}.
\newblock \emph{IEEE Transactions on Robotics}, 33\penalty0 (6):\penalty0
  1273--1291, 2017.
\newblock \doi{10.1109/TRO.2017.2721939}.

\bibitem[Singh(1989)]{cd_definition1989singh}
A.~Singh.
\newblock \href{https://doi.org/10.1080/01431168908903939}{Review Article
  Digital change detection techniques using remotely-sensed data}.
\newblock \emph{International Journal of Remote Sensing}, 10\penalty0
  (6):\penalty0 989--1003, 1989.

\bibitem[Kenney et~al.(2009)Kenney, Buckley, and Brock]{kenney2009interactive}
J.~Kenney, T.~Buckley, and O.~Brock.
\newblock \href{https://ieeexplore.ieee.org/document/5152393}{Interactive
  Segmentation for Manipulation in Unstructured Environments}.
\newblock In \emph{Proceedings of the 2009 IEEE International Conference on
  Robotics and Automation}, ICRA{\rq}09, pages 1343--1348. IEEE Press, 2009.

\bibitem[{Elgammal} et~al.(2002){Elgammal}, {Duraiswami}, {Harwood}, and
  {Davis}]{rw_backgroundsubkde2002elgammal}
A.~{Elgammal}, R.~{Duraiswami}, D.~{Harwood}, and L.~S. {Davis}.
\newblock \href{https://ieeexplore.ieee.org/document/1032799}{Background and
  foreground modeling using nonparametric kernel density estimation for visual
  surveillance}.
\newblock \emph{Proceedings of the IEEE}, 90\penalty0 (7):\penalty0 1151--1163,
  July 2002.

\bibitem[Bouwmans et~al.(2019)Bouwmans, Javed, Sultana, and
  Jung]{bouwmans2018deep}
T.~Bouwmans, S.~Javed, M.~Sultana, and S.~K. Jung.
\newblock
  \href{http://www.sciencedirect.com/science/article/pii/S0893608019301303}{Deep
  Neural Network Concepts for Background Subtraction: {A} Systematic Review and
  Comparative Evaluation}.
\newblock \emph{Neural Networks}, 117:\penalty0 8--66, 2019.

\bibitem[{Jianbo Shi} and {Malik}(1998)]{shi1998motionsegm}
{Jianbo Shi} and J.~{Malik}.
\newblock \href{https://ieeexplore.ieee.org/document/710861}{Motion
  segmentation and tracking using normalized cuts}.
\newblock In \emph{IEEE International Conference on Computer Vision (ICCV)},
  pages 1154--1160, Jan 1998.

\bibitem[Horn and Schunck(1981)]{of_of1981Horn}
B.~K. Horn and B.~G. Schunck.
\newblock Determining optical flow.
\newblock \emph{Artificial Intelligence}, 17\penalty0 (1):\penalty0 185--203,
  1981.

\bibitem[Tu et~al.(2019)Tu, Xie, Zhang, Poppe, Veltkamp, Li, and
  Yuan]{zhigang2019survey}
Z.~Tu, W.~Xie, D.~Zhang, R.~Poppe, R.~C. Veltkamp, B.~Li, and J.~Yuan.
\newblock A survey of variational and cnn-based optical flow techniques.
\newblock \emph{Signal Processing: Image Communication}, 72:\penalty0 9--24,
  2019.

\bibitem[Fragkiadaki et~al.(2015)Fragkiadaki, Arbelaez, Felsen, and
  Malik]{fragkiadaki2015learning}
K.~Fragkiadaki, P.~Arbelaez, P.~Felsen, and J.~Malik.
\newblock \href{https://arxiv.org/abs/1412.6504}{Learning to segment moving
  objects in videos}.
\newblock In \emph{Proc. {IEEE} Conf. on Computer Vision and Pattern
  Recognition (CVPR)}, pages 4083--4090, 2015.

\bibitem[Bideau et~al.(2018)Bideau, RoyChowdhury, Menon, and
  Learned-Miller]{bideau2018best}
P.~Bideau, A.~RoyChowdhury, R.~R. Menon, and E.~Learned-Miller.
\newblock
  \href{http://openaccess.thecvf.com/content\_cvpr\_2018/html/Bideau\_The\_Best\_of\_CVPR\_2018\_paper.html}{The
  Best of Both Worlds: Combining CNNs and Geometric Constraints for
  Hierarchical Motion Segmentation}.
\newblock In \emph{The IEEE Conference on Computer Vision and Pattern
  Recognition (CVPR)}, June 2018.

\bibitem[Dave et~al.(2019)Dave, Tokmakov, and Ramanan]{dave2019towards}
A.~Dave, P.~Tokmakov, and D.~Ramanan.
\newblock \href{http://arxiv.org/abs/1902.03715}{Towards Segmenting Everything
  That Moves}.
\newblock \emph{CoRR}, abs/1902.03715, 2019.

\bibitem[Xie et~al.(2018)Xie, Xiang, Fox, and Harchaoui]{xie2018objdiscovery}
C.~Xie, Y.~Xiang, D.~Fox, and Z.~Harchaoui.
\newblock \href{http://arxiv.org/abs/1812.02772}{Object Discovery in Videos as
  Foreground Motion Clustering}.
\newblock \emph{CoRR}, abs/1812.02772, 2018.

\bibitem[Shao et~al.(2018)Shao, Shah, Dwaracherla, and
  Bohg]{robot_motionsegmentation2018shao}
L.~Shao, P.~Shah, V.~Dwaracherla, and J.~Bohg.
\newblock \href{http://arxiv.org/abs/1804.05195}{Motion-based Object
  Segmentation based on Dense {RGB-D} Scene Flow}.
\newblock \emph{CoRR}, abs/1804.05195, 2018.

\bibitem[Fitzpatrick and Metta(2003)]{robot_motionsegmentation2003fitzpatrick}
P.~Fitzpatrick and G.~Metta.
\newblock \href{https://doi.org/10.1098/rsta.2003.1251}{Grounding vision
  through experimental manipulation}.
\newblock \emph{Philosophical Transactions of the Royal Society of London.
  Series A: Mathematical, Physical and Engineering Sciences}, 361\penalty0
  (1811):\penalty0 2165--2185, Aug. 2003.

\bibitem[{Schiebener} et~al.(2014){Schiebener}, {Ude}, and
  {Asfour}]{robot_motionsegmentation2014schiebener}
D.~{Schiebener}, A.~{Ude}, and T.~{Asfour}.
\newblock \href{https://ieeexplore.ieee.org/document/6907586}{Physical
  interaction for segmentation of unknown textured and non-textured rigid
  objects}.
\newblock In \emph{2014 IEEE International Conference on Robotics and
  Automation (ICRA)}, pages 4959--4966, May 2014.

\bibitem[{Schiebener} et~al.(2011){Schiebener}, {Ude}, {Morimoto}, {Asfour},
  and {Dillmann}]{schiebener2011segandlearning}
D.~{Schiebener}, A.~{Ude}, J.~{Morimoto}, T.~{Asfour}, and R.~{Dillmann}.
\newblock \href{https://ieeexplore.ieee.org/document/6100843}{Segmentation and
  learning of unknown objects through physical interaction}.
\newblock In \emph{2011 11th IEEE-RAS International Conference on Humanoid
  Robots}, pages 500--506, Oct 2011.

\bibitem[Pathak et~al.(2018)Pathak, Shentu, Chen, Agrawal, Darrell, Levine, and
  Malik]{pathak2018learning}
D.~Pathak, Y.~Shentu, D.~Chen, P.~Agrawal, T.~Darrell, S.~Levine, and J.~Malik.
\newblock
  \href{https://people.eecs.berkeley.edu/~pathak/papers/cvprw18.pdf}{Learning
  Instance Segmentation by Interaction}.
\newblock In \emph{CVPR Workshop on Benchmarks for Deep Learning in Robotic
  Vision}, 2018.

\bibitem[Du et~al.(2019)Du, Wang, and Lian]{rw_visiongraspingsurvey2019du}
G.~Du, K.~Wang, and S.~Lian.
\newblock
  \href{https://dblp.org/rec/bib/journals/corr/abs-1905-06658}{Vision-based
  Robotic Grasping from Object Localization, Pose Estimation, Grasp Detection
  to Motion Planning: {A} Review}.
\newblock \emph{CoRR}, abs/1905.06658, 2019.

\bibitem[d.~{Rocha} et~al.(2019)d.~{Rocha}, {Padoy}, and {Rosa}]{rocha2019self}
C.~C. d.~{Rocha}, N.~{Padoy}, and B.~{Rosa}.
\newblock \href{https://arxiv.org/abs/1902.04810}{Self-Supervised Surgical Tool
  Segmentation using Kinematic Information}.
\newblock In \emph{2019 International Conference on Robotics and Automation
  (ICRA)}, pages 8720--8726, 2019.

\bibitem[Florence et~al.(2019)Florence, Corso, and Griffin]{florence2019self}
V.~Florence, J.~J. Corso, and B.~Griffin.
\newblock \href{http://arxiv.org/abs/1904.00952}{Self-Supervised Robot In-hand
  Object Learning}.
\newblock \emph{CoRR}, abs/1904.00952, 2019.

\bibitem[{Otsu}(1979)]{otsu1979threshold}
N.~{Otsu}.
\newblock \href{https://ieeexplore.ieee.org/document/4310076}{A Threshold
  Selection Method from Gray-Level Histograms}.
\newblock \emph{IEEE Transactions on Systems, Man, and Cybernetics}, 9\penalty0
  (1):\penalty0 62--66, Jan 1979.

\bibitem[Lin et~al.(2014)Lin, Maire, Belongie, Bourdev, Girshick, Hays, Perona,
  Ramanan, Doll{\'a}r, and Zitnick]{dataset_mscoco2014lin}
T.~Lin, M.~Maire, S.~J. Belongie, L.~D. Bourdev, R.~B. Girshick, J.~Hays,
  P.~Perona, D.~Ramanan, P.~Doll{\'a}r, and C.~L. Zitnick.
\newblock \href{http://arxiv.org/abs/1405.0312}{Microsoft {COCO:} Common
  Objects in Context}.
\newblock \emph{CoRR}, abs/1405.0312, 2014.

\bibitem[Chen et~al.(2018)Chen, Zhu, Papandreou, Schroff, and
  Adam]{chen2018encoder}
L.-C. Chen, Y.~Zhu, G.~Papandreou, F.~Schroff, and H.~Adam.
\newblock \href{https://arxiv.org/abs/1802.02611}{Encoder-Decoder with Atrous
  Separable Convolution for Semantic Image Segmentation}, 2018.

\bibitem[Gonz{\'a}lez(2009)]{gonzales2009measurement}
{\'A}.~Gonz{\'a}lez.
\newblock \href{http://dx.doi.org/10.1007/s11004-009-9257-x}{Measurement of
  Areas on a Sphere Using Fibonacci and Latitude--Longitude Lattices}.
\newblock \emph{Mathematical Geosciences}, 42\penalty0 (1):\penalty0 49--64,
  Nov 2009.

\bibitem[Mai et~al.(2017)Mai, Gouiff{\`e}s, and Bouchafa]{mai2017refinement}
T.~K. Mai, M.~Gouiff{\`e}s, and S.~Bouchafa.
\newblock
  \href{https://www.semanticscholar.org/paper/Optical-Flow-Refinement-using-Reliable-Flow-Mai-Gouiff\%C3\%A8s/9f742b85e3373119ec232edf98bea083230278d8}{Optical
  Flow Refinement using Reliable Flow Propagation}.
\newblock In \emph{VISIGRAPP}, 2017.

\bibitem[Xiang et~al.(2017)Xiang, Schmidt, Narayanan, and
  Fox]{methods_ycbvideodataset2017xiang}
Y.~Xiang, T.~Schmidt, V.~Narayanan, and D.~Fox.
\newblock \href{http://arxiv.org/abs/1711.00199}{PoseCNN: {A} Convolutional
  Neural Network for 6D Object Pose Estimation in Cluttered Scenes}.
\newblock \emph{CoRR}, abs/1711.00199, 2017.

\bibitem[Shi and Koonjul(2014)]{methods_autgrasp2014shi}
J.~Shi and G.~S. Koonjul.
\newblock
  \href{https://www.researchgate.net/publication/285406656\_Real-time\_Grasp\_Planning\_with\_Environment\_Constraints}{Real-time
  Grasp Planning with Environment Constraints}.
\newblock In \emph{IROS 2014 Workshop: Real-time Motion Generation \& Control},
  2014.

\bibitem[{Radke} et~al.(2005){Radke}, {Andra}, {Al-Kofahi}, and
  {Roysam}]{cd_survey2005radke}
R.~J. {Radke}, S.~{Andra}, O.~{Al-Kofahi}, and B.~{Roysam}.
\newblock \href{https://ieeexplore.ieee.org/document/1395984}{Image change
  detection algorithms: a systematic survey}.
\newblock \emph{IEEE Transactions on Image Processing}, 14\penalty0
  (3):\penalty0 294--307, 2005.

\bibitem[He et~al.(2017)He, Gkioxari, Doll{\'a}r, and Girshick]{he2017mask}
K.~He, G.~Gkioxari, P.~Doll{\'a}r, and R.~Girshick.
\newblock \href{https://arxiv.org/abs/1703.06870}{Mask {R}-{CNN}}.
\newblock \emph{arXiv preprint arXiv:1703.06870}, 2017.

\bibitem[Agisoft(2018)]{agisoft2018agisoft}
L.~Agisoft.
\newblock
  \href{https://www.agisoft.com/pdf/metashape-pro\_1\_5\_en.pdf}{Agisoft
  metashape user manual, Professional edition, Version 1.5}.
\newblock \emph{Agisoft LLC, St. Petersburg, Russia, accessed June}, 2018.

\bibitem[Maji(2011)]{is_turk_pvoc2011maji}
S.~Maji.
\newblock
  \href{http://www2.eecs.berkeley.edu/Pubs/TechRpts/2011/EECS-2011-79.html}{Large
  Scale Image Annotations on Amazon Mechanical Turk}.
\newblock Technical Report UCB/EECS-2011-79, EECS Department, University of
  California, Berkeley, Jul 2011.

\end{thebibliography}
